\documentclass[journal]{IEEEtran}
\usepackage[english]{babel}
\usepackage[utf8]{inputenc}
\usepackage{amsmath}
\usepackage{mathptmx}
\usepackage{graphicx}
\usepackage[font=small, skip=1 pt]{caption}
\usepackage{cite}
\usepackage{makecell}
\usepackage{csvsimple}
\usepackage{pbox}
\usepackage{tabularx,booktabs}
\usepackage{float}
\usepackage{multirow}
\usepackage{makecell}
\usepackage{algcompatible}
\usepackage{algorithm}
\usepackage{mathtools}
\usepackage{xcolor}
\newcommand\norm[1]{\left\lVert#1\right\rVert}
\ifCLASSINFOpdf
\else
 \fi
\begin{document}
\title{Intelligent Condition Based Monitoring Techniques for  Bearing Fault Diagnosis}
\author{Vikas Singh,~\IEEEmembership{Student Member,~IEEE,}
            and~   Nishchal K. Verma,~\IEEEmembership{Senior Member,~IEEE}
  \thanks{Vikas Singh,   and Nishchal K. Verma are with the Dept. of Electrical Engineering, Indian Institute of Technology, Kanpur, India, 208016.
e-mail: (vikkyk@iitk.ac.in,   nishchal@iitk.ac.in).}}
\markboth{}%
{Shell \MakeLowercase{\textit{et al.}}: Bare Demo of IEEEtran.cls for IEEE Journals}
\maketitle
\begin{abstract}
In recent years, intelligent condition-based monitoring of rotary machinery systems has become a major research focus of machine fault diagnosis. In condition-based monitoring, it is challenging to form a large-scale well-annotated dataset due to the expense of data acquisition and costly annotation. The generated data have a large number of redundant features which degraded the performance of the machine learning models. To overcome this, we have utilized the advantages of minimum redundancy maximum relevance (\textit{mRMR}) and transfer learning with a deep learning model. In this work, \textit{mRMR} is combined with deep learning and deep transfer learning framework to improve the fault diagnostics performance in terms of accuracy and computational complexity. The \textit{mRMR} reduces the redundant information from data and increases the deep learning performance, whereas transfer learning, reduces a large amount of data dependency for training the model. In the proposed work, two frameworks, i.e., \textit{mRMR} with deep learning and \textit{mRMR} with deep transfer learning, have explored and validated on CWRU and IMS rolling element bearings datasets. The analysis shows that the proposed frameworks can obtain better diagnostic accuracy compared to existing methods and can handle the data with a large number of features more quickly.

\end{abstract}
\IEEEpeerreviewmaketitle

\section{Introduction}
\label{intro}

\IEEEPARstart{W}{ith} the recent advancement of technology, intelligent condition monitoring of rotating machines become an essential tool of machine fault diagnosis to increase the reliability and ensure the equipment efficiency in industrial processes \cite{c1,c2,c3}. Rotating components, which are essential parts of machines, are widely used in equipment transmission systems, and their failure might result in considerable loss and catastrophic consequences. As practical components for condition-based maintenance, the vibration-based fault diagnosis systems have been explored in recent years \cite{c4}.

Traditionally, machine fault diagnosis framework includes three main stages: 1) signal acquisition, 2) feature extraction and selection, 3) fault identification or classification. The signal acquisition stage involves the collection of raw data while the machine is in running condition. The signals such as vibration,  temperature, current,  sound pressure, and acoustic emission can be studied for health monitoring and fault diagnosis, but the vibration signal is extensively explored in the literature because it provides essential information about the faults  \cite{c5,c6,c7}.  In the second stage, feature extraction is used to extract informative features from the raw data using time-domain, frequency-domain, and time-frequency domain analysis \cite{c8}. Although these feature extraction methods identify the machine health conditions, however, they may have  irrelevant and insensitive features which affect the fault diagnosis performance. Therefore, feature selection methods such as mutual information: criteria of max-dependency, max-relevance, and min-redundancy, principal component analysis (PCA) and Fisher discriminant analysis (FDA) is widely used to select the essential features from the data \cite{c9, c10,c11}. In the final stage, selected features are used for fault classification using various classifier, i.e., support vector machine (SVM), \textit{k}-nearest neighbor (\textit{k}-NN), random forest (RF),  and artificial neural network (ANN) \cite{c12}. However, there are also several limitations with these traditional fault diagnosis methods. 
 
  In  2006,  Hinton \textit{et al.} \cite{c13} proposed deep learning techniques which trying to learn the high-level representation of data by stacking the multiple layers in the hierarchical architecture. In recent years, several studies have focused on deep neural network (DNN) for machine fault diagnosis.  Tao \textit{et al.} \cite{c14} suggested a deep neural network framework for bearing fault diagnosis based on stacked auto-encoder and softmax regression. In \cite{c15,maurya} authors have proposed a DNN-based intelligent fault diagnosis method for the classification of different datasets from bearings element  and gearboxes with large samples using auto-encoder.     Sun \textit{et al.} \cite{c16}, proposed a sparse auto-encoder-based DNN with the help of the denoising coding and dropout method for induction motor fault diagnosis. Ding \textit{et al.} \cite{c18}  developed a deep convnet in which wavelet packet energy has used as input for bearing fault diagnosis. In \cite{c19}, intelligent machine bearings fault diagnosis method is presented by combining the  compressed data acquisition and deep learning approach in a single framework.
 
 Although deep learning-based models have achieved great success in machine fault diagnosis applications, however, there are still problems that are associated with deep learning models. As the number of hidden layers and nodes is increasing, the number of parameters also increased, which increases the computational complexity of the model. Along with that large amount of labeled training data is required for training the deep network from scratch. In addition to that, parameter optimization and hyperparameter tuning greatly affect the performance of deep networks. Transfer learning-based approaches have been used to overcome these problems where instead of training the deep learning models from scratch, a DNN that has been trained on sufficient labelled training data in different running conditions is used and fine-tuned on the target task. In the literature, various case studies have been performed  using transfer learning with the DNN.  Lu \textit{et al.}  \cite{tr1}  proposed a DNN model with domain adaptation for fault diagnosis. First, they utilize the domain adaptation to strengthen the representation of the original data, so that high classification accuracy can be obtained on the target domain. Second, they proposed various strategies to explore the optimal hyperparameters of the model.  Long \textit{et al.}\cite{tr2} have presented a deep transfer learning-based model using sparse auto-encoder (SAE) for fault diagnosis. In which three layers SAE is used to extract the abstract features of raw data and uses the maximum mean discrepancy term to minimize the discrepancy between the features of training and testing data. Gao \textit{et al.} \cite{tr3}  presented an intelligent fault diagnosis of machines with unlabeled data using deep convolutional transfer learning network.  Siyu \textit{et al.} \cite{tr4} have proposed a highly accurate machine fault diagnosis model using deep transfer learning with a convolution neural network. However, the performance of these models again reduced due to redundant features in the dataset because, in the presence of redundant features, the different initial condition will lead to different performance.

In this paper, we have addressed the problem mentioned above, by employing the mutual information: criteria of max-dependency, max-relevance, and min-redundancy to select a subset of features with minimum redundancy. The selected features are used to train the DNN to extract meaningful representation in the lower dimension. In this work, two frameworks of intelligent fault diagnosis method have been evaluated. In the first framework, the DNN is pre-trained and fine-tuned on the same running condition, and the fine-tuned network is validated on unseen samples of same running conditions.  However, in the second framework,  the deep neural network is pre-trained on one running condition with unlabelled data and pre-trained weights are transferred on target domain and finally target network is fine-tuned on different running condition with less number of sample. So in the second framework, the pre-training time of the target network is totally eliminated.  In the real application, as mentioned, it is challenging to form a large-scale well-annotated dataset. In this scenario, the second framework is more useful.

The major contributions of the paper are summarized as:
\begin{enumerate}
    \item   \textit{mRMR} with deep learning and \textit{mRMR} with deep transfer learning frameworks have been proposed for an intelligent machine fault diagnosis, as shown in Fig 2.
    \item \textit{mRMR} based machine learning-based method has been evaluated to minimize the effect of redundant features in the dataset. The redundant features decrease the performance of the deep models because in the presence of redundant features, the different initial condition will lead to different performance.
    \item Deep learning and deep transfer learning based methods have been evaluated for better feature representation in reduced dimension with lower complexity.
    \item Confusion matrix chart is used  to describe the performance of the classification model.
    \item Experiments have been conducted on CWRU \cite{dataset} and IMS \cite{ims} datasets to show the efficacy of the proposed approach in comparison of state-of-the-art methods.
\end{enumerate}

The remainder of this paper is organized as follows: Section \ref{theoreticalbc} briefly introduces the theoretical background for \textit{mRMR}, deep neural network and transfer learning. Section \ref{proposed} describes the proposed \textit{mRMR}-DNN based transfer learning framework for intelligent condition based monitoring. Section \ref{experi} presents the experimental results and analysis of proposed method in comparison with state-of-the-art methods. Finally, Section \ref{conclusion} draw the conclusion of complete paper. 
\section{Theoretical Background}
\label{theoreticalbc}
\subsection{Minimum Redundancy Maximum Relevance (\textit{mRMR})}
The mutual information is used to determine the feature set $S$ with $m$ features which jointly have the maximum dependency on the target class $y$ \cite{mRMR}. This approach is termed as Max-Dependency and written as 
\begin{align}
\max\, D(S,y), \,\, D = I (\{x_i ,\, i=1, \cdots, m\}; \,y)
\label{maxd}
\end{align}
where, $I$ is the mutual information.

The Max-Dependency criterion is difficult to implement in the high-dimensional feature space: 1) the number of samples is often insufficient and 2) multivariate density estimation often involves computing the inverse of high-dimension covariance matrix,  which is generally an ill-posed. This problem is overcome by maximal relevance criterion (Max-Relevance). Max-Relevance is used to find features satisfying (\ref{maxr}), which approximates $D(S, y)$ in (\ref{maxd}) with the mean values of all mutual information values between individual features $x_i$ and class $y$:
\begin{align}
\max \, D(S,\,y), \,\, D = \frac{1}{|S|} \sum_{x_i \in S}I(x_{i};\, y)
\label{maxr}
\end{align}

Features selected using Max-Relevance could likely have sufficient redundancy, i.e., the dependency among these features could be significant. When two features highly depend on each other, the respective class-discriminative power would not change much if one of them were eliminated. Therefore, the following minimal redundancy (Min-Redundancy) condition can be used to select mutually exclusive features as
\begin{align}
\min \, R(S,\,y), \,\, R = \frac{1}{|S|} \sum_{x_i \in S}I(x_{i},\, x_{j})
\label{minr}
\end{align}

By combining the above two criteria, i.e., (\ref{maxr}) and (\ref{minr}), it is called “minimal-redundancy-maximal-relevance” (\textit{mRMR}).  The operator $\phi (D, R)$  is used to combine $D$ and $R$ and considered as the simplest form to optimize $D$ and $R$ simultaneously:
\begin{align}
\label{optimize}
\max\, \phi (D,\, R), \,\, \phi = D-R  
\end{align}
\subsection{Deep Learning (DL)}
\begin{figure}
	\centering
	\includegraphics[width=0.7\linewidth]{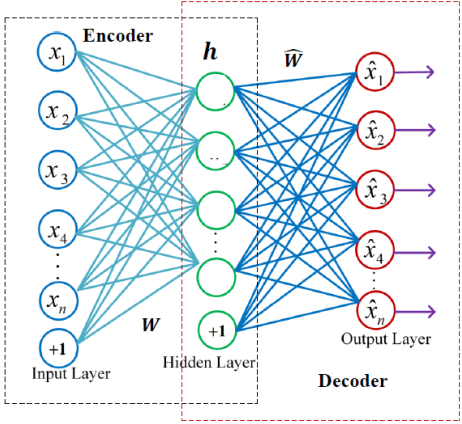}
	\caption{Basic architecture of sparse auto-encoder.}
    \label{Fig: autoencoder}
\end{figure}

Deep learning is a branch of machine learning and its fundamental principle to learn a hierarchical representation of data from layer to layer \cite{c13}. In the literature, a different type of deep learning models has been studied for machine fault diagnosis. In this paper, we have used a sparse auto-encoder based learning approach to form a deep neural network for automatic feature extraction. 
The auto-encoder is a three-layer feed-forward neural network comprises an input layer, hidden layer, and output layer. As shown in figure \ref{Fig: autoencoder} the first part is known as an encoder which takes input $ \textbf{x} $ and transforms it into a hidden representation $\textbf{h}$ via a non-linear mapping as 
\begin{align}
    \textbf{h} = f(\textbf{W} \textbf{x} +\textbf{b})
    \label{encoder}
\end{align}

where, $f$ is a non-linear activation function. The second part of the figure is known as a decoder which maps the hidden representation $\textbf{h}$ back to original representation as
\begin{align}
    \hat{ \textbf{x} } = f(\hat{ \textbf{W}} \textbf{h} + \hat{ \textbf{b}})
    \label{decoder}
\end{align}

\begin{figure*}[!ht]
	\centering
    	\includegraphics[width=15cm]{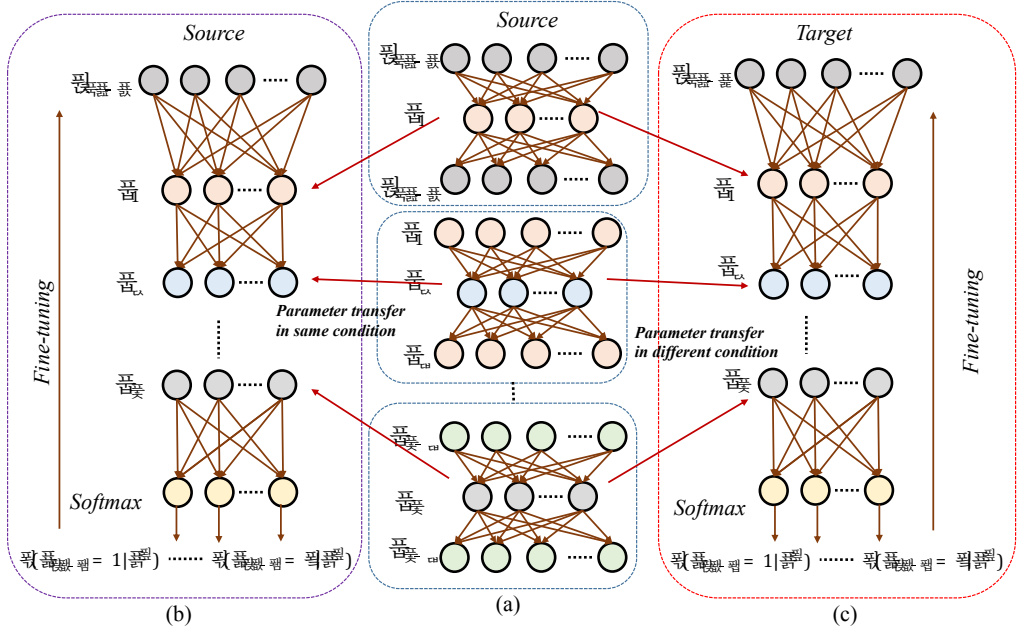}
  		\caption{Proposed Framework: (a) Pre-training of sparse auto-encoder  (b) Training of deep neural network with labelled data are available in a source domain   (c)  Transferring the pre-trained weight of source data on target network with small sample are available in target domain.}
    \label{Fig: DNN_TL}
\end{figure*}
The network parameters, i.e., weight (\textbf{W}) and bias (\textbf{b}), are optimized by minimizing the cost function $J_{sparse}(\textbf{W, b})$ using the back-propagation by computing the gradients with respect to  \textbf{W} and  \textbf{b}. 
\begin{align}
\begin{split}
J_{sparse}(\textbf{W, b})=\frac{1}{n}\sum_{i=1}^{n}\frac{1}{2}\norm{\textbf{x}_i
- \hat{\textbf{x}_i}}^{2}\\+\frac{\lambda}{2} \norm{\textbf{W}}^{2}+\beta\sum_{j=1}^{s}KL(\rho\lVert\hat \rho_{j})
\end{split}
\end{align}

where, $KL$ (Kullback- Leibler) is the divergence function, $s$ is the number of hidden nodes, $\lambda$ is regularization parameter, $\rho$ is sparsity parameter and $\beta$ is a sparsity control parameter. The  $\hat{\rho}_j$ is the average activation of hidden node $j$.
\subsection{Transfer Learning (TL)}
Transfer learning \cite{TL} is a form of representation learning based on idea of mastering a new task by reusing knowledge from a previous task and it is defined as:

Given a source domain $D_S$ and learning task $T_S$, a target domain $D_T$ and a learning task $T_T$, TL tries to improve the learning performance of target function $f_T (\cdot)$ in $D_T$ using the knowledge in $D_S$ and $T_S$, where $D_S \neq D_T$, or $T_S \neq T_T$. Based on the source task and target task it is categorized in three subcategories:
\begin{enumerate}
\item Inductive TL: Given a source domain $D_S$ and a learning task $T_S$, a target domain $D_T$ and a learning task $T_T$, inductive TL tries to improve the learning performance of target predictive $f_T (\cdot)$ in $D_T$ using the knowledge in $D_S$ and $T_S$, where $T_S \neq T_T$.
\item Transductive TL: Given a source domain $D_S$ and a learning task $T_S$, a target  domain $D_T$ and a  learning task $T_T$, transductive TL tries to improve the learning performance of  target function $f_T (\cdot)$ in $D_T$ using the knowledge in $D_S$ and $T_S$, where $D_S \neq D_T$ and $T_S = T_T$ . 
\item Unsupervised TL: Given a source domain $D_S$ with a learning task $T_S$, a target domain $D_T$ and a learning task $T_T$, unsupervised TL aims to improve the learning of the target function $f_T (\cdot)$  in $D_T$ using the knowledge in $D_S$ and $T_S$, where $T_S \neq T_T$ and $Y_S$ and $Y_T$ are not observable.
\end{enumerate}

In this work, we have utilized the inductive transfer learning to evaluate the performance on CWRU and IMS rolling element bearings datasets.

\section{Proposed frameworks for Fault Diagnosis}
\label{proposed}
Considering the challenges posed by traditional fault diagnosis  methods in the condition-based monitoring system, this paper presents an intelligent condition-based monitoring framework by minimizing the redundant features from data and transferring the knowledge from one domain to different domain. In this work, \textit{mRMR} based feature selection method is utilized to eliminate the effect of redundant features from the dataset as described in Algorithm 1. The redundant features decrease the performance of the deep learning models because, in the presence of redundant features, different initial condition will lead to different performance. The data with reduced features are utilized to pre-train the source network as shown in figure \ref{Fig: DNN_TL}a, in the case of deep neural network the pre-trained model is fine-tuned on the source data as shown in figure \ref{Fig: DNN_TL}b, and validated on the unseen target data with same machine running condition as described in Algorithm 2. However, in the case of deep transfer learning,  the DNN with  inductive transfer learning are applied where the target task is different from the source task, no matter when the source and target domains are the same or different. In inductive transfer learning setting, we have evaluated a condition that  a lot of labeled training data are available in the source domain and small labeled training data available in the target domain, as illustrated in figure \ref{Fig: DNN_TL}c.  

As shown in figure \ref{Fig: DNN_TL}a,  sparse auto-encodes are learned layer by layer in an unsupervised way on the source data. The sparse auto-encoder learned at $l^{th}$ is given as 

\begin{align}
\begin{split}
J(\textbf{W, b})=\frac{1}{n}\sum_{i=1}^{n}\frac{1}{2}\norm{\textbf{h}_i^{l}
-\hat{\textbf{h}_i}^{l}}^{2}\\+\frac{\lambda}{2}\sum_{i=1}^{s_{l}}\sum_{j=1}^{s_{l+1}}\textbf{W}_{ji}^{2}+\beta\sum_{j=1}^{s_{l}}KL(\rho\lVert\hat \rho_{j})
\end{split}
\end{align} 

where $\textbf{h}_i^{l}$ and $\hat{\textbf{h}_i}^{l}$ are the hidden input and and estimated hidden output of the $l^{th}$ sparse auto-encoder, $s_l$ and $s_{l+1}$ are number of nodes in the $\textbf{h}_i^{l}$ and $\hat{\textbf{h}_i}^{l}$,  $\rho$ is the sparsity and $KL$ is the Kullback-Leibler.  The sparsity and $KL$ divergence at activation ($a_j^{l}$) of hidden unit $j$ are defined as follows: 
\begin{align} 
\rho_j=\frac{1}{m}\sum_{i=1}^{m} a_j^{l}(h_i) \\
\sum_{j=1}^{s_{l}} KL(\rho||\rho_j) =\sum_{j=1}^{s_{l}} \rho\log\frac{\rho}{\rho_j}+(1-\rho)\log\frac{1-\rho}{1-\rho_j} 
\end{align}

The encoding output (i.e., $z^1, z^2, \cdots, z^n$, where, $i=1, 2, \cdots, n$) of last  SAE is used as input to the softmax layer for pre-training of the softmax layer and is given as 
\begin{gather}
 h_{\theta}(z^i) 
 = \begin{bmatrix}
   p(y^{i}=1 |z^i; \theta) \\
   p(y^{i}=2 |z^i; \theta) \\
   \vdots\\
   p(y^{i}=k |z^i; \theta) \\
   \end{bmatrix}
   =
 \frac{1}{\sum_{\mu =1}^ke^{\theta_{\mu}^Tz^i}} 
  \begin{bmatrix}
   e^{\theta_{1}^Tz^i}\\
   e^{\theta_{2}^Tz^i}\\
   \vdots\\
  e^{\theta_{k}^Tz^i}\\
   \end{bmatrix}
\end{gather}

The learned sparse auto-encoders are stacked with the softmax layer to form the deep neural network as shown in figure \ref{Fig: DNN_TL}b. The deep network shown in figure \ref{Fig: DNN_TL}b is fine-tuned on the source label to obtain the optimal weight and bias vectors of the network as defined below. \begin{align}
J_{DNN}(\theta_s) = \text{Loss} (y_s, \hat{y}_s)
\end{align} 

The parameter (i.e., weight and bias)  of the pre-trained model as shown figure \ref{Fig: DNN_TL}a are transferred on target network figure \ref{Fig: DNN_TL}c and are work as initial parameter for the target domain. The obtained optimal network has trained on the target domain that has less labeled data as described in Algorithm \ref{DTL} and given as follow: \begin{align}
J_{DTL} (\theta_t)= \text{Loss}(y_t, \hat{y}_t)
\end{align} 

The trained model on the target domain is validated on unseen test data of the target domain.

\begin{algorithm*}

\caption{Relevant Feature Selection and Pre-training of Sparse Auto-encoder}
\label{FSAE}
\begin{algorithmic}[1]
\item $\textbf{x} \longleftarrow data$,  $\textbf{y} \longleftarrow class \,\,output$ \hspace{190pt}  \# \textbf{x} and  \textbf{y} are the input and output vectors
\item $[Initial\,\,features,\, Selected \,\, features, \,\textbf{y}] \leftarrow Input_{\textit{mRMR}}$   \#  initial number of features, selected number of features and class $\textbf{y}$
\item $\text{for}\,\, i\longleftarrow 1\,\, : \,\,  number\,\, of\,\, initial\,\, features$
\item $relevance \longleftarrow mutualInfo(x_{i},\, y) $ \hspace{33pt} \#  find the relevance between individual features $x_i$ and  class $y$ as defined in (\ref{maxr})
\item $\text{for}\,\, j\longleftarrow 1\,\, : \,\,  number\,\, of\,\, initial\,\, features$
\item $redundancy \longleftarrow mutualInfo(x_{i},\, x_{j}) $  \hspace{15pt}  \# find the redundancy between individual features $x_i$ and  class $x_j$ as defined in (\ref{minr})
\item $ mRMRValues(x_{i}) \longleftarrow relevance - redundancy $  \hspace{65pt}  \# compute the difference between relevance and redundancy
\item $selectedFeatures \longleftarrow sort(mRMRValues) $ \hspace{10pt}  \# sort the features on the basis of minimum redundancy maximum relevance
\item $Input \longleftarrow selectedFeatures$ \hspace{160pt}  \# input with minimum redundant features in the data
\item $SAE \longleftarrow [Input, HiddenLayer_{1}, Input]$ \hspace{160pt} \# learning of $1^{st}$ sparse auto-encoder (SAE)
\item $SAE_{l} \longleftarrow [HiddenLayer_{l-1}, HiddenLayer_{l}, HiddenLayer_{l-1}]$ \hspace{60pt}  \# similarly learning of $l^{th}$ sparse auto-encoder
\end{algorithmic}
\end{algorithm*}
\begin{algorithm*}
\caption{Fine-tuning of Deep Neural Network (DNN)}
\label{FTDNN}
\begin{algorithmic}[1]
\item $SSAE \longleftarrow [Input, HiddenLayer_{1}, HiddenLayer_{2}, ..., HiddenLayer_{l}]$\,\,\,\,\,\,\,\,\,\, \# stacked all hidden layer with input to form SSAE
\item  $DNN \longleftarrow [SSAE, Softmax\,\, layer]$  \hspace{30pt}  \# form DNN with help of stacked sparse auto-encoder  (SSAE) and softmax layer
\item $J(W, b) \longleftarrow mean \, square \, error\,(Y_{RC-A}, \hat{Y}_{RC-A})$  \hspace{5pt}  \#  use actual labels ($Y_{RC-A}$) of the source data  and estimated labels ($\hat{Y}_{RC-A}$) of the source data
\item $Finetuned \, network \longleftarrow backpropagation$ \hspace{5pt}  \# the whole network is fine-tuned on the source labels using back-propagation to find the optimal parameter, i.e., weight and bias vectors
\item $trained\, DNN \longleftarrow Finetuned \, network$  \hspace{140pt} \# the fine-tuned network is the trained DNN model
\item $Predicted\, output \longleftarrow trained\, DNN$ \hspace{133pt} \# test the trained DNN on unseen source data RC-A 
\end{algorithmic}
\end{algorithm*}
\begin{algorithm*}
\caption{Deep Transfer Learning (DTL)}
\label{DTL}
\begin{algorithmic}[1]
\item $SSAE \longleftarrow [Input, HiddenLayer_{1}, ..., HiddenLayer_{l}]$ \hspace{35pt} \# Pre-trained SSAE on source data of step 1 in Algorithm 2 
\item $Softmax \,\, layer \longleftarrow pretraining [SSAE_n, \hat{Y}_{RC-B}] $  \hspace{15pt} \# pre-training of softmax layer on target labels using final output of the last SSAE
\item  $DNN \longleftarrow [SSAE, Softmax\,\, layer]$  \hspace{150pt}  \# form DNN with help of SSAE and softmax layer 
\item $J(W, b) \longleftarrow mean \, square \, error\,(Y_{RC-B}, \hat{Y}_{RC-B})$  \hspace{5pt}  \#  use labels ($Y_{RC-B}$) of the target data RC-B and estimated labels ($\hat{Y}_{RC-B}$)
\item $Finetuned \, network \longleftarrow backpropagation$ \hspace{5pt}  \# the whole network is fine-tuned on the target labels using back-propagation to find the optimal parameter, i.e., weight and bias vectors
\item $trained\, DNN \longleftarrow Finetuned \, network$  \hspace{140pt} \# the fine-tuned network is the trained DNN model
\item $Predicted\, output \longleftarrow trained\, DNN$ \hspace{150pt} \# test the trained DNN on unseen target data RC-B 
\end{algorithmic}
\end{algorithm*}
\begin{table*}[ht]
\centering %
\caption{\textsc{  \small Dataset Description }} %
\begin{tabular}{ |c|c|c|c|c|c|c|c|c|c|}
\hline
    &  \multicolumn{4}{c|} {Case 1: CWRU data with same fault diameter} & \multicolumn{2}{c|} {Case 2: Different  fault diameter} & \multicolumn{2}{c|} {Case 3: IMS Data} & \\
\cline{2-9}
 \makecell{ Class}  &  \multicolumn{2}{c|} {Source: RC-A} & \multicolumn{2}{c|} {Target: RC-B}  &  \multicolumn{2}{c|} {Target: RC-B} & \multicolumn{2}{c|} {Target: RC-B} & label \\
\cline{2-9}
 \multirow{2}{*} &   Number of Sample & Load & Number of Sample & Load &   Number of Sample & Load &  Number of Sample & Load & \\
\cline{1-10}
 Normal & 1210  & 0 hp  & 400, 400, 400 & 1, 2, 3 &  400, 400, 400  &1, 2, 3& 400 &26.6 kN  &1  \\
\cline{1-10}
 Inner & 1210   & 0 hp& 400, 400, 400  & 1, 2, 3 &  400, 400, 400  &1, 2, 3 &400 & 26.6 kN  &2 \\
\cline{1-10}
 Ball & 1210   & 0 hp& 400, 400, 400  & 1, 2, 3 & 400, 400, 400 &1, 2, 3&400  & 26.6 kN  & 3 \\
\cline{1-10}
 Outer & 1210   &0 hp & 400, 400, 400  & 1, 2, 3 & 400, 400, 400  &1, 2, 3 &400  & 26.6 kN  &4 \\
\cline{1-10}
\end{tabular}
\label{Table: dataset description}
\end{table*}
\begin{table}[ht]
\centering %
\caption{\textsc{  \small RC-A is a source data and  RC-B is a target data }} %
\begin{tabular}{ |c|c|c|}
\hline
   Conditions  & Source (RC-A) & Target (RC-B)  \\
\cline{1-3}
 Normal-inner race  &  $2420 \times 100$  & $800 \times 100$  \\ \cline{1-3}
 Normal-outer race  &  $2420 \times 100$  & $800 \times 100$ \\ \cline{1-3}
  Normal-outer race  &  $2420 \times 100$  & $800 \times 100$ \\ \cline{1-3}
  Normal-inner-ball-outer  &  $4840 \times 100$  & $1600\times 100$  \\ 
\hline
\end{tabular}
\label{Table: Case 1}
\end{table}
\section{Experimental Results and Analysis}
\label{experi}
The proposed frameworks are validated on two different case studies, i.e., Case Western Reserve University (CWRU) Bearing Data \cite{dataset} and  Intelligent Maintenance Systems (IMS) Bearing Data \cite{ims}. They are described as follows:
\subsection{Dataset Description}
Experimental data are taken from the CWRU and IMS data center to analyze the performance of the proposed frameworks. Experimental setup of CWRU and IMS bearing test rig has shown in figures \ref{Fig: Apparatus} and \ref{Fig: Apparatus1}  by which multivariate vibration series has generated for the validation. The CWRU test stand consists of a 2-hp Reliance Electric motor on the left of stand, a torque transducer/encoder in the center, a dynamometer on the right, and control electronics are not shown in the figure. Single point faults with diameters of 7, 14 and 21 mils ((1 mil = 0.001 inches) have seeded at the inner raceway, rolling element, and outer raceway of the test bearing using electro-discharge machining. The vibration data are collected using accelerometers and driver end vibration signal have used, which have 12 kHz (12,000 samples per second) sampling rate with 2 hp load. However, the IMS, data are collected at 20 kHz sampling rate with 26.6 kN load.

In this analysis, dataset included four health conditions: 1) normal condition, 2) outer race fault, 3) inner race fault, and 4) roller fault with two fault diameters 7 and 14 mils.

\begin{table*}[ht]
\centering %
\caption{\textsc{  \small  Accuracy  with Softmax Classifier on Different Running Condition with Same Fault Diameters (7 mils)}} %
\begin{tabular}{ |c|c|c|c|c|c|c|c|c|c|c|c|}
\hline
    \multirow{2}{*}{Dataset}  &  \multirow{2}{*} {\makecell{Motor\\ Load}}  & \multirow{2}{*}{Condition} & \multirow{2}{*}{\makecell{PCA \\ \cite{c10}}} &  \multirow{2}{*}{\makecell{RFE \\ \cite{svm-rfe}}} &  \multirow{2}{*}{ \makecell{\textit{mRMR}\\ \cite{mRMR}}}&  \multicolumn{3}{c|} {DNN} & \multicolumn{3}{c|} {Without Source Label}  \\
\cline{7-12}
    & & & & & & DANN \cite{dann}&DNN \cite{sdnn}  &  \makecell{DNN \\with \textit{mRMR}} &NDTL \cite{tr2}  &  DTL  & \makecell{DTL\\ with \textit{mRMR}}\\
\cline{1-12}
 \multirow{9}{*}{Binary-Class} & 1 &  \multirow{3}{*}{Normal-ball}& 82.75   &  78.00  &79.00 &89.23 & 99.36 &      \textbf{99.36}& 99.20 &      98.75 &         \textbf{99.00}  \\
\cline{2-2} \cline{4-12}
& 2 &      & 84.25  &71.75 &71.75 &91.67  & 99.63 &  \textbf{ 99.88} & 92.40    &     99.00  &      \textbf{99.00 } \\
\cline{2-2} \cline{4-12}
&3 &      & 89.00  &71.75   &78.50 & 98.23 & 99.38  &      99.00 & 99.10   &    99.36  &      \textbf{99.75}  \\
 \cline{2-12}
&  1 &  \multirow{3}{*}{Normal-inner}&    68.50   &   57.75   	&  61.00  & 92.40 & 99.75 &         \textbf{99.75} &  99.50   &     99.25 &      \textbf{99.50}  \\
\cline{2-2} \cline{4-12}
& 2 &       &  73.50&69.00 & 72.25 & 92.86 & 99.50 &      \textbf{99.50} &  92.38   &     99.00 &      \textbf{ 99.75}  \\
\cline{2-2} \cline{4-12}
&3 &  &  55.50   & 58.00  & 61.75 & 92.84  & 100.00 &      \textbf{100.00} &  98.90  &   99.63 &         99.00   \\
 \cline{2-12}
&  1 &  \multirow{3}{*}{Normal-outer}& 81.50   & 71.25  &79.00  &99.70 & 100.00 &     \textbf{ 100.00} &  100.00&       100.00 &        \textbf{100.00} \\
\cline{2-2} \cline{4-12}
& 2 &      & 57.50  &61.25  & 66.00 &  98.60& 100.00 &        \textbf{100.00} & 100.00   &     100.00 &        \textbf{100.00}  \\
\cline{2-2} \cline{4-12}
&3 & &79.00    & 69.00  & 72.50 & 98.10 & 100.00 &        \textbf{100.00} & 98.85 &    100.00 &        \textbf{100.00} \\
\hline
 \multirow{3}{*}{Multi-Class} & 1 &  \multirow{3}{*}{\makecell{Normal-inner- \\ outer-ball}}&38.12   & 77.50 &  47.62 &78.44 & 85.44 &        \textbf{ 86.56} & 89.50 &     86.15 &         \textbf{86.50}  \\
\cline{2-2} \cline{4-12}
& 2 &    &26.38     & 80.63 & 45.13  & 72.19  & 88.88 &      \textbf{ 89.88}     & 82.38 &  88.69 &    \textbf{ 89.56} \\
\cline{2-2} \cline{4-12}
&3 & & 30.12      & 80.63 &  45.13  & 83.44  & 92.06 &      91.44 & 89.88  &  90.69 &     90.19 \\
\hline
\end{tabular}
\label{Table: DRCSFD}
\end{table*}
\begin{table*}[ht]
\centering %
\caption{\textsc{  \small  Accuracy  with Softmax Classifier on Different Running Condition with Fault Diameters (14 mils)}} %
\begin{tabular}{ |c|c|c|c|c|c|c|c|c|c|c|c|}
\hline
    \multirow{2}{*}{Dataset}  &  \multirow{2}{*} {\makecell{Motor\\ Load}}  & \multirow{2}{*}{Condition} & \multirow{2}{*}{\makecell{PCA \\ \cite{c10}}} &  \multirow{2}{*}{\makecell{RFE \\ \cite{svm-rfe}}} &  \multirow{2}{*}{ \makecell{\textit{mRMR}\\ \cite{mRMR}}}&  \multicolumn{3}{c|} {DNN} & \multicolumn{3}{c|} {Without Source Label}  \\
\cline{7-12}
    & & & & & & DANN \cite{dann} &DNN \cite{sdnn} &  \makecell{DNN \\ with \textit{mRMR}} & NDTL \cite{tr2}  &  DTL  & \makecell{DTL \\with \textit{mRMR}}\\
\cline{1-12}
 \multirow{9}{*}{Binary-Class} & 1 &  \multirow{3}{*}{Normal-ball}&83.50    & 97.75  & 79.50 &82.20 & 99.25 &      \textbf{99.50} &  91.00   &  98.50 &     \textbf{ 98.75}   \\
\cline{2-2} \cline{4-12}
& 2 &    &  92.50     & 88.25 & 88.50 & 82.60 & 96.63 &      \textbf{96.88} & 95.20    &  96.75 &     \textbf{ 96.88}   \\
\cline{2-2} \cline{4-12}
&3 & & 90.75   & 84.50 & 88.75 &77.00 & 96.00 &      \textbf{96.88} &   96.30  &  96.50 &      96.13  \\
 \cline{2-12}
&  1 &  \multirow{3}{*}{Normal-inner}&79.50    & 82.00 & 79.00 &77.30 & 97.25 &      \textbf{97.50} &  97.75  &   98.00 &      97.50    \\
\cline{2-2} \cline{4-12}
& 2 &    &  79.75    & 83.00 & 74.25 &76.90  & 96.38 &      \textbf{96.38} &  96.25  &  96.75 &      \textbf{96.75}  \\
\cline{2-2} \cline{4-12}
&3 & & 87.75  & 81.25 & 78.75&72.30  &  98.38 &      98.25 &95.81  &     97.50 &      \textbf{98.00}  \\
 \cline{2-12}
&  1 &  \multirow{3}{*}{Normal-outer}&  95.25    & 75.50 & 96.00  & 81.40& 97.75 &      \textbf{98.63}     & 97.00&  97.25 &      \textbf{97.25}   \\
\cline{2-2} \cline{4-12}
& 2 &    &89.50    & 79.75 & 82.25 & 83.43  & 97.75 &      \textbf{97.75} &  97.50    & 97.50 &      \textbf{97.50}    \\
\cline{2-2} \cline{4-12}
&3 & &  89.75  & 84.75 &  86.75 & 87.88 & 97.13 &      \textbf{98.25} & 95.50   &   94.63 &      \textbf{95.50}   \\
\hline
 \multirow{3}{*}{Multi-Class} & 1 &  \multirow{3}{*}{\makecell{Normal-inner- \\ outer-ball}}& 70.00  & 78.13 & 71.62 & 33.10 & 95.63 &     \textbf{96.44} &   71.00  &  95.00 &      \textbf{95.00 }  \\
\cline{2-2} \cline{4-12}
& 2 &    &   63.12  & 79.37 & 70.00 &20.91 & 86.19 &      \textbf{86.44} &  65.20  &   87.63 &     \textbf{87.69}   \\
\cline{2-2} \cline{4-12}
&3 & &   74.38   & 81.25 &   71.12 & 31.90& 91.19 &      89.81 & 67.30  &     90.44 &     \textbf{92.00} \\
\hline
\end{tabular}
\label{Table: DRCDFD}
\end{table*}
\begin{table*}[ht]
\centering %
\caption{\textsc{  \small  Accuracy  with Softmax Classifier on Intelligent Maintenance System Dataset}} %
\begin{tabular}{ |c|c|c|c|c|c|c|c|c|c|c|c|}
\hline
    \multirow{2}{*}{Dataset}  &  \multirow{2}{*} {\makecell{Motor\\ Load}}  & \multirow{2}{*}{Condition} & \multirow{2}{*}{\makecell{PCA \\ \cite{c10}}} &  \multirow{2}{*}{\makecell{RFE \\ \cite{svm-rfe}}} &  \multirow{2}{*}{ \makecell{\textit{mRMR}\\ \cite{mRMR}}}&  \multicolumn{3}{c|} {DNN} & \multicolumn{3}{c|} {Without Source Label}  \\
\cline{7-12}
    & & & & & & DANN \cite{dann} & DNN \cite{sdnn}  &  \makecell{DNN \\with \textit{mRMR}} &NDTL \cite{tr2}  &  DTL  & \makecell{DTL\\ with \textit{mRMR}}\\
\cline{1-12}
 \multirow{3}{*}{Binary-Class} &\multirow{4}{*}{26.6} &  \multirow{1}{*}{Normal-ball}&53.00   & 47.00  & 51.75 & 71.13 &72.34 &	\textbf{72.72} & 81.14 & 	80.59 &	\textbf{82.67} \\                                                                                  \cline{3-12} 
&   &  Normal-inner   &  45.25   & 41.0 & 50.50 & 80.23 & 84.72	& \textbf{86.84}	 & 88.62 & 89.66	& \textbf{90.44} \\
\cline{3-12}
& & Normal-outer & 98.25  & 99.15 & 99.25 & 98.82 & 99.38 &	\textbf{99.38} & 98.91 & 	99.38 &	\textbf{99.50} \\
\cline{1-1} \cline{3-12}
 \multirow{1}{*}{Multi-Class} & &  \makecell{Normal-inner-\\outer-ball}&48.38    & 46.25  & 47.25 &73.63 & 73.14 &	\textbf{74.72} & 72.94 & 	74.81	& \textbf{75.08}  \\
 \hline
\end{tabular}
\label{Table: IMS}
\end{table*}
\subsection{Segmentation}
The length of time series data is  massive (at least 121,000 samples in CWRU and 20,000 sample in IMS), if all the sample directly applied to machine learning algorithms, it will take ample time to train the network.  To overcome this problem the training samples for all conditions have been segmented with the size of one quarter of the sampling period to learn the local characteristics of the signal \cite{segmen}. Three cases have  been investigated for the analysis of proposed frameworks as described below and given in Table  \ref{Table: dataset description}.
\begin{enumerate}
\item  Case 1: In this case, running condition A (RC-A) with fault diameter 7 mils has been treated as a source data, where each type of data have at least 121,000 samples with 12kHz frequency and approximately 1797 RPM motor speed. Therefore the number of sample points per revolution is around 400. The number of sample points has been taken one-quarter of the sampling frequency, so the total number of sample points for each type of data is 1210, and the dimension of each sample points is 100.  The running condition B (RC- B) with same diameter (7 mils) and different load (1hp, 2hp, 3hp) has been treated as target data where each type of data have at least 40,000 sample points with same frequency and speed. The total number of sample points for each type of data is 400, and the dimension of each sample is 100. 
\item Case 2:  In this case, running condition A (RC-A) with fault diameter, 7 mils has been treated as a  source data as similar to case 1.  However, running condition B  (RC-B) is changed here fault diameter 14 mils with different load  (1hp,  2hp,  3hp)  has treated as target data where each type of data have at least 40,000 sample points with 12kHz frequency and 1797 RPM motor speed. The total number  of  sample  points  for  each  type  of  data  is  400, and the dimension of each sample points is 100. 
\item Case 3:  In this case, running condition A (RC-A) with fault diameter, 7 mils has treated as a  source data as similar to case 1 and 2. However, running condition B \begin{figure}[H]
	\centering
	\includegraphics[width=0.95 \linewidth]{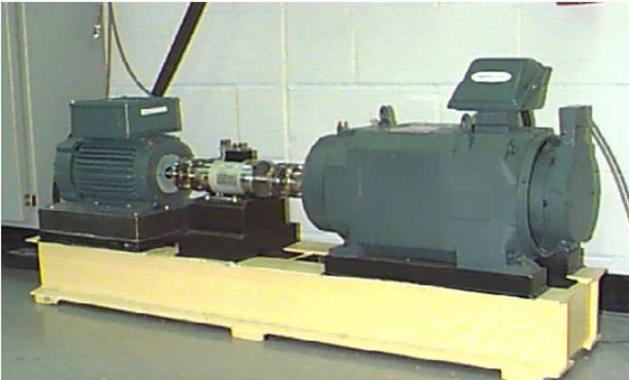}
	\caption{Apparatus of CWRU bearing  used for the experiment \cite{dataset}.}
    \label{Fig: Apparatus}
\end{figure} (RC-B) is changed here by IMS data with load 26.6 kN  has treated as target data where each type of data have at least 40,000 sample points (with two different data files) with 20 kHz frequency and 2000 RPM  speed. The total number  of  sample  points  for  each  type  of  data  is taken  as 400, and the dimension of each sample points is 100.  In this case the knowledge learned from one machine (CWRU) is transferred to different machine (IMS).  \begin{figure*}[h]
    \centering
	\includegraphics[width=18cm]{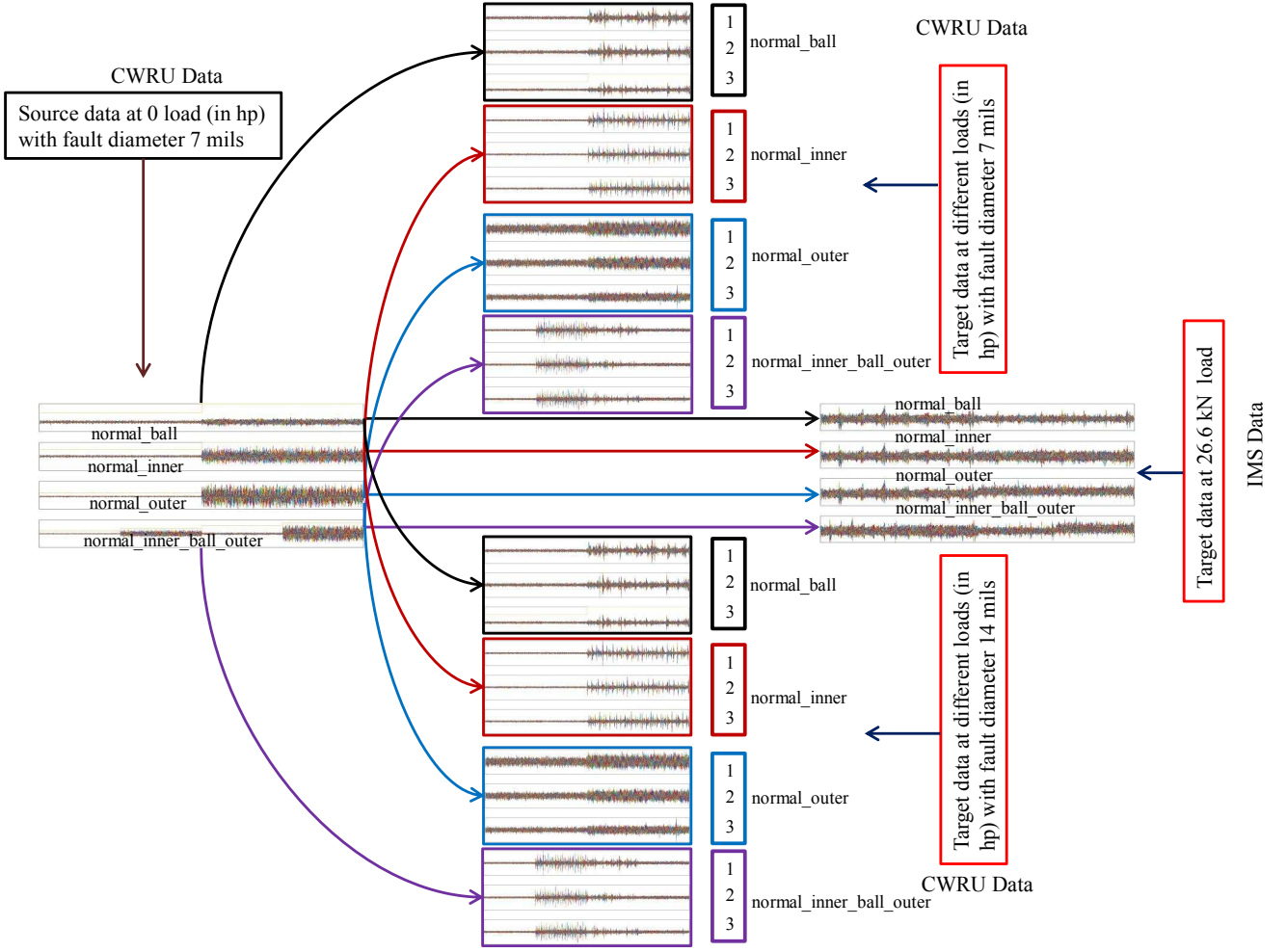}
	\caption{Transfer of knowledge from source data to different target data in case of DTL and DTL with \textit{mRMR}.}
    \label{Fig: flow diagram}
    \end{figure*}\begin{figure}[H]
	\centering
	\includegraphics[width=0.75\linewidth]{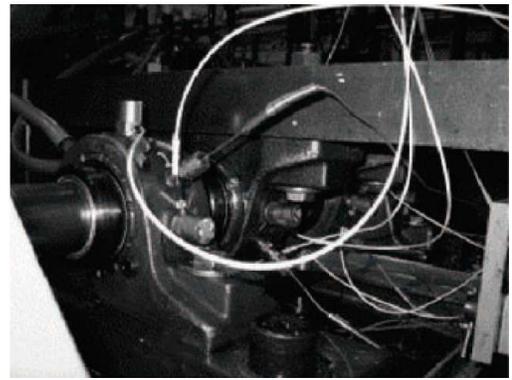}
	\caption{Apparatus of  IMS bearing used for the experiment \cite{ims}.}
    \label{Fig: Apparatus1}
    \end{figure}  
       
\end{enumerate}

\subsection{Pre-processing}
The data collected from the accelerometers are not well structured. If  the network is trained on such type of dataset, they perform poorly. So to make it well structured, they are pre-processed before training the network. In this paper, max-min normalization is used to evaluate the performance.
\begin{align}\label{normalized}
\textbf{x}_{norm}=\frac{\textbf{x}-\textbf{x}_{min}}{\textbf{x}_{max}-\textbf{x}_{min}}
\end{align}

where, $\textbf{x}$ is the un-normalized data, $\textbf{x}_{norm}$ is the normalized data,  $\textbf{x}_{min}$ and $\textbf{x}_{max}$ are minimum and maximum values of the data.  The pre-processed data is sampled using  5-fold external cross-validation  for  better generalization of the model.
\begin{figure*}
\centering %
	\includegraphics[width=17cm]{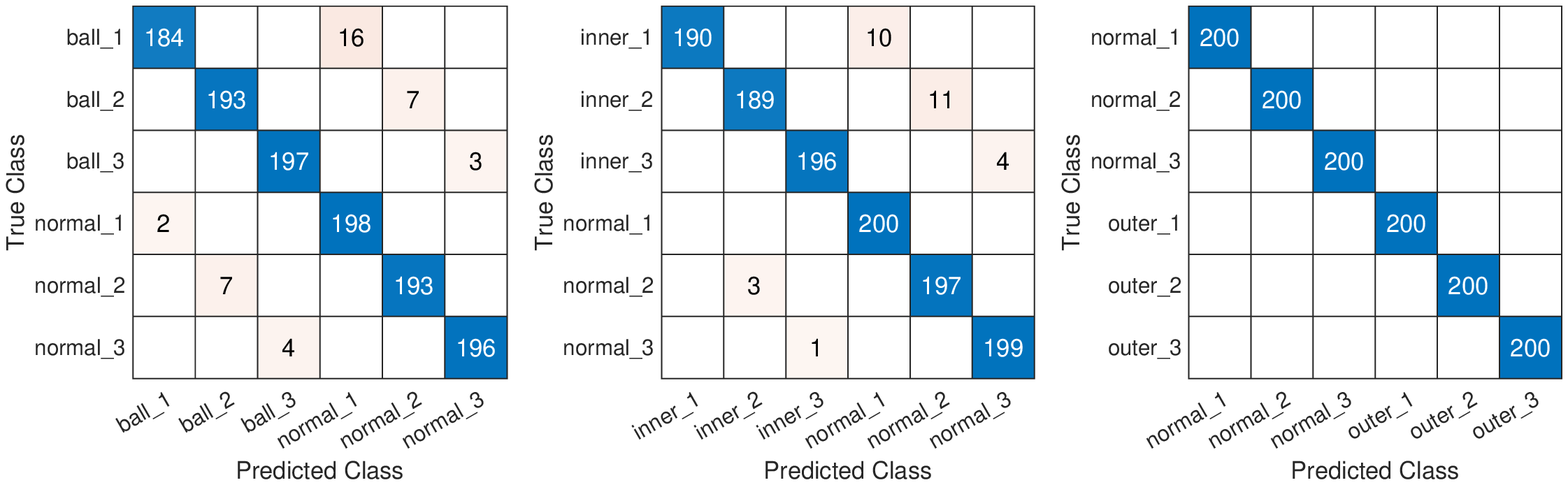}
	\caption{Confusion matrix plots of the predicted results for DNN with \textit{mRMR}}
    \label{Fig: confusion1}
\end{figure*}
\begin{figure*}
\centering %
	\includegraphics[width=17cm]{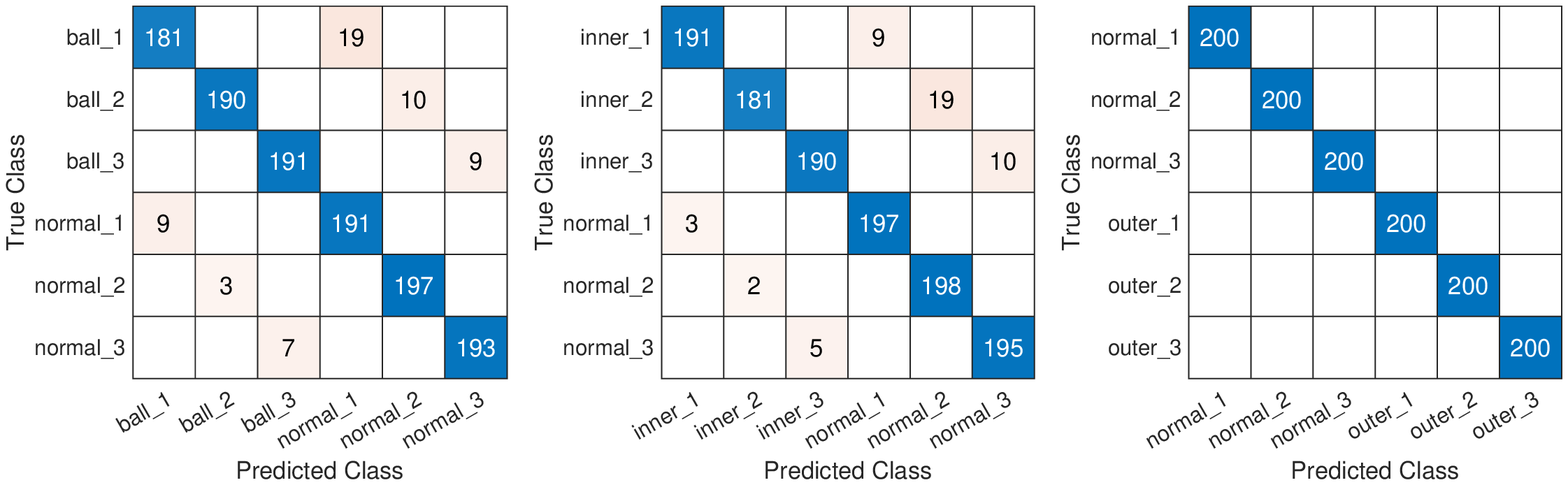}
	\caption{Confusion matrix plots of the predicted results for DTL with \textit{mRMR}}
    \label{Fig: confusion}
\end{figure*}
\subsection{Analysis}
In this subsection,  Case 1, Case 2 and Case 3  as described in Table \ref{Table: dataset description} are analyzed on binary-class and multi-class classification problem  (as described in Table \ref{Table: Case 1}) and compared with DNN and DTL without \textit{mRMR} and state-of-the-art feature selection/extraction methods.
\subsubsection{Comparison with DNN and DTL without \textit{mRMR}}
The performance of proposed frameworks are compared with deep learning and deep transfer learning without reducing the effect of redundant features using \textit{mRMR} on a binary and multi-class classification problem. As given in Table \ref{Table: dataset description}, the source data is considered at load 0 with the normal condition and load 0 with fault diameter  7 mils with other conditions (i.e., inner, ball, outer). The target domain is considered at different load with the same diameter (7 mils), different diameters (14 mils) and at different machine (IMS) as given in Table \ref{Table: dataset description} and described in Case 1, Case 2 and Case 3. The network architecture chosen in DNN is $100\times 50 \times 40 \times 20$, where, 100 is input layer, 50 is first hidden layer, 40 is second hidden layer, and 20 is the last hidden layer with regularization $\lambda=10^{-3}$   and  $\beta =0.3$. The sparsity parameter $\rho$ is varied from \{0.1, 0.2, 0.3, 0.4, 0.5, 0.6, 0.7, 0.8, 0.9\} and result is reported on the best sparsity parameter. Whereas in DNN with \textit{mRMR} firstly, $70$ features are selected with minimum redundancy then a DNN is formed with network architecture $70\times 50 \times 40 \times 20$. The transformed features at the last layer of DNN and DNN with \textit{mRMR} are applied to the softmax classifier to identify the health state of the rotatory machines.  

In DTL, the pre-trained DNN weight parameters on source data are used as initial weight of the target network. The target network is fine-tuned on the target label, and the fine-tuned network is validated on the unseen target data. Whereas, in DTL with \textit{mRMR},  the pre-trained DNN with \textit{mRMR} weight parameters on source data are used as an initial weight of the target network.  So in the transfer learning model, no pre-training is required, which reduces the training time of the model.  The performance of the model, i.e., DNN with \textit{mRMR} and DTL with \textit{mRMR} are compared in the term of accuracy.  The features extracted using the proposed models are applied to the softmax classifier to interpret the performance in terms of accuracy values. As given in Table \ref{Table: DRCSFD},  \ref{Table: DRCDFD} and \ref{Table: IMS}, the prediction accuracy on the unseen data are better and comparable. The confusion matrix charts are also presented in figures \ref{Fig: confusion1} and \ref{Fig: confusion} for the case of DNN with \textit{mRMR} and DTL with \textit{mRMR} between the actual health state vs predicted health state to describe the performance of the classification model.

\begin{table}[b]
\centering %
\caption{\textsc{  \small Comparison of execution times (in seconds) }} %
\begin{tabular}{ |c|c|c|}
\hline
  S. No. & Methods & \makecell{Execution time \\
}   \\
\cline{1-3}
1& PCA + Softmax Classifier  & $2.62 + 38.81$  \\ \cline{1-3}
 2& SVM-RFE + Softmax Classifier  & $1.80 +38.81$ \\ \cline{1-3}
3& \textit{mRMR}+ Softmax Classifier  &   $4.40 + 38.81$ \\ \cline{1-3}
4&  DANN + Softmax Classifier &  $5443.74 + 38.81$  \\ \cline{1-3}
5&  DNN + Softmax Classifier &  $5.28 + 38.81$  \\ \cline{1-3}
 5&  DNN-\textit{mRMR} + Softmax Classifier  &  $\textbf{4.12} + 38.81$  \\ \cline{1-3}
 6&  NDTL  + Softmax Classifier &  $799.79 + 38.81$  \\ \cline{1-3}
 7&  DTL + Softmax Classifier & $3.78+38.81$  \\ \cline{1-3}
  8&    DTL-\textit{mRMR} + Softmax Classifier   &   $\textbf{2.83} + 38.81$  \\ \cline{1-3}
\hline
\end{tabular}
\label{Table: Execution time}
\end{table}

\subsubsection{Comparison  with Traditional State-of-the-art Feature Selection/Extraction Methods}
In this subsection, the performance of proposed frameworks are compared with traditional methods, i.e., PCA\cite{c10}, \textit{mRMR} \cite{mRMR} and support vector machine recursive feature elimination (SVM-RFE) \cite{svm-rfe} on a binary and multi-class classification problem.  In PCA  features are extracted in accordance of highest to lowest eigenvalues,  \textit{mRMR} features are selected as explained in the subsection \ref{theoreticalbc}-A and described in Algorithm 1 whereas, in SVM-RFE the features are selected based on ranking in which top-ranked features are chosen. In all these state-of-the-art methods same numbers of features are chosen as of proposed frameworks. As given show  in Table \ref{Table: DRCSFD}, \ref{Table: DRCDFD} and \ref{Table: IMS}, the performance of the proposed frameworks is better in term of accuracy values.  The execution time of the proposed approaches are also compared with  these methods as shown in \ref{Table: Execution time}. The execution time of DNN-\textit{mRMR} is less in compared to DNN, whereas, for DTL and DTL-\textit{mRMR}, the execution time reduced  because in the case of DTL no pre-training of the network are required from the scratch as described in the proposed approach.

\section{Conclusion}
\label{conclusion}
This paper presents a new framework for intelligent machine fault diagnosis. The major contributions of this paper are to minimize the effect of redundant features in the dataset and transfer the knowledge to a different domain. Because, in the presence of redundant features, the different initial conditions will lead to different performances. However, knowledge transfer helps to improve the performance of the target domain with less number of sample in the target domain. To overcome this effect, \textit{mRMR} feature selection is utilized to reduce the redundant features, and data with reduced redundant features are used for training the DNN and pre-trained model on different running conditions is fine-tuned on the target task. The proposed frameworks are validated on the famous motor bearing dataset from CWRU and IMS data center. The results in terms of accuracy values show that the proposed frameworks are an effective tool for machine fault diagnosis. The effectiveness is also represented in terms of  confusion matrix chart. In the future, the proposed frameworks are very helpful in the application area like, bioinformatics, where the number of features is very large in comparison to the number of samples. The proposed approach will help to reduce the non-informative features from such type of data and improve the performance.

\bibliographystyle{IEEEtran.bst}
\bibliography{Reference.bib}

\end{document}